\documentclass{article}

\PassOptionsToPackage{numbers, compress}{natbib}

\usepackage[preprint]{neurips_2019}

\usepackage[utf8]{inputenc} 
\usepackage[T1]{fontenc}    
\usepackage{hyperref}       
\usepackage{url}            
\usepackage{booktabs}       
\usepackage{amsfonts}       
\usepackage{amsmath}
\usepackage{nicefrac}       
\usepackage{microtype}      
\usepackage{graphicx}
\usepackage{subfig}
\usepackage{algorithm}
\usepackage[noend]{algpseudocode}
\usepackage{float}
\usepackage{adjustbox}
\usepackage{makeidx}
\usepackage{multirow}
\usepackage{multicol}
\usepackage{bbm}

\DeclareMathOperator*{\argmax}{arg\,max}
\DeclareMathOperator*{\argmin}{arg\,min}

\title{Dynamic Regularizer with an Informative Prior}

\author{%
  Avinash Kori*\\
  Department of Engineering Design\\
  Indian Institute of Technology, Madras\\
  \texttt{avinashgkori@smail.iitm.ac.in} \\
  \And
   Manik Sharma\thanks{equal contribution} \\
   Department of Engineering Design\\
   Indian Institute of Technology, Madras\\
   \texttt{ed15b024@smail.iitm.ac.in} \\
}

\begin{document}

\maketitle

\begin{abstract}
Regularization methods, specifically those which directly alter weights like $L_1$ and $L_2$, are an integral part of many learning algorithms. Both the regularizers mentioned above are formulated by assuming certain priors in the parameter space and these assumptions, in some cases, induce sparsity in the parameter space. Regularizers help in transferring beliefs one has on the dataset or the parameter space by introducing adequate terms in the loss function. Any kind of formulation represents a specific set of beliefs: $L_1$ regularization conveys that the parameter space should be sparse whereas $L_2$ regularization conveys that the parameter space should be bounded and continuous. These regularizers in turn leverage certain priors to express these inherent beliefs. A better understanding of how the prior affects the behavior of the parameters and how the priors can be updated based on the dataset can contribute greatly in improving the generalization capabilities of a function estimator. In this work, we introduce a weakly informative prior and then further extend it to an informative prior in order to formulate a regularization penalty, which shows better results in terms of inducing sparsity experimentally, when compared to regularizers based only on Gaussian and Laplacian priors. Experimentally, we verify that a regularizer based on an adapted prior improves the generalization capabilities of any network.  We illustrate the performance of the proposed method on the MNIST and CIFAR-10 datasets.

\end{abstract}

\section{Introduction}

A large number of metrics \cite{largemetric1, largemetric2, largemetric3} have been proposed over the years which aim to gauge the magnitude of a vector. Most of these approaches are in the Euclidean space and have been bunched together in the form of $L_p$ norms. Though each of these norm functions are formulated in a way most suitable for the required objective, they have been used in varied forms in the deep learning community, often serving as losses and at times, as regularizers. Hence, the motivation behind each regularization is quite different. Regularizers which directly alter the parameters of the function estimator utilize these functions heavily. In the last few decades, people have tried to incorporate many additional constraints, along with these learning algorithms, to enforce smoothness \cite{reg_smoothness} and sparsity. For instance, the use of $L_1$ regularizer enforces sparsity and the use of $L_2$ regularizer enforces bound on the magnitude and introduces continuity in the parameter space. 

Another important dimension which has been explored is controlling the sparsity of the network. This approach follows the Bayesian paradigm, in which priors providing more information about the parameter space, as compared to the traditional ones, are used. In our case, the prior tries to accentuate the part of the parameter space which is significant and masks that part of the parameter space which is insignificant. \cite{group_sparse_reg} analyzed the effect of group-level sparsity on deep neural network and proposed a generalized approach for optimizing network weights, network architecture and feature selections. \cite{benifits_group_sparsity} discussed the advantages of strong group sparsity using the theory of group lasso and have also provided theoretical justification for using group sparse regularizer. \cite{bayesian_compression} discussed a method to optimize the model by enforcing the constraints on the sparsity of the network and achieve state of the art results in terms of compression of the model.

In Bayesian statistical inference, a prior (represented in the form of a probability distribution) is a device which is used to express one’s belief about a quantity before any kind of evidence is considered. Using the Bayes’ theorem, we can calculate the posterior probability distribution which takes into account the evidence (data). When definite information about a variable is known (for e.g, the empirical expectation and variance through previous experiments), it can be expressed using an informative prior. On the other hand, if only a little is known about the range of the variable, it can be expressed in the form of a weakly informative prior. The third class of priors, called \textit{uninformative priors}, are the ones which are used to lay equal bets over all the plausible outcomes, for instance, if the quantity under consideration is positive or within a limit.

In this work, we propose a new form of regularizer based on informative priors. We make use of a prior which dynamically adapts to the type of weights during training. The prior is constantly updated as the probability distribution of the parameters keeps changing. The parameters, in turn, are updated by sampling this probability distribution. To compensate the abrupt shift in the distribution, we dampen its effect by introducing a momentum term which restricts the sudden changes in the parameters by combining the previous and current instances of the parameters in a convex manner. We evaluate our results on MNIST handwritten digit dataset and CIFAR10 natural images dataset. Further sections in this paper describe our methodology and results in greater detail.

\section{Methodology}

\subsection{MAP estimate of an informative prior :}
\label{method}    
We can investigate the formulation of the regularizer by assuming our estimation function to be a linear regressor. The regressor is parameterized by a vector, $W = \{w_i\}$.

\[\hat{y} = w_0 x_0 + w_1x_1 + w_2 x_2 + . . . + w_N x_N  = \sum_{i=0} ^N w_i x_i \]

The probability distribution of the parameter space characterized by W can be formulated in the Bayesian paradigm as shown in Eqn. \ref{bayesian_formulation}. Using this formulation, it is possible to localize a point in the parameter space by maximizing the overall probability on the left-hand side.
\begin{equation}
    \mathbb{P}(W|X) = \frac{\mathbb{P}(X|W) \mathbb{P}(W)} {\mathbb{P}(X)}
    \label{bayesian_formulation}
\end{equation}


\begin{equation}
 W_{MAP} = \argmax_W \log (\mathbb{P}(X|W)) + \log (\mathbb{P}(W))
   \label{split_eqn}
\end{equation}

Taking the negative log-likelihood on both sides splits the right side of the Eq. \ref{split_eqn} in two terms. One term is responsible for learning, and the other term is responsible for enforcing our beliefs about the parameters. This, in effect, improves the generalization ability of the model. The resulting form is the maximum a-posteriori estimate of the parameters. 

We assume a prior where the current instance of the parameter is sampled from a probability distribution provided by the previous state of the parameter space. The probability distribution is a mixture of a softmax and a uniform distribution which, due to the threshold factor ($T$), results in a multiple-spike and slab kind of distribution which is strictly controlled by the dynamically changing parameters. The prior depends on the distribution of weights, which makes it an informative prior and explicit in comparison to Gaussian or Laplacian priors. 




\[ W _{MAP} = \argmax_W \left[\log \prod_{i=1}^n \frac{1}{\sigma \sqrt{2\pi}} e ^ {-\frac{(y_i - (w_0 x_{i, 0} + . . . + w_N x_{i, N} ))^2} {2\sigma^2}} + \log \prod_{j = 0}^N \frac{1}{\tau \sqrt{2 \pi}} e ^{-\frac{\sum_S (w_j \mathbbm{1}(\mathbb{P}_s (w_j) > T) )^2}{2\tau^2}} \right]\]

\[ = \argmax_W \left[\log \prod_{i=1}^n \frac{1}{\sigma \sqrt{2\pi}} e ^ {-\frac{(y_i - \hat{y_i})^2}{2\sigma^2}} + \log \prod_{j = 0}^N \frac{1}{\tau \sqrt{2 \pi}} e ^{-\frac{\sum_S (w_j \mathbbm{1}(\mathbb{P}_s (w_j) > T) )^2}{2\tau^2}} \right]\]

\[ = \argmax_W \left[\sum_{i=1}^n \log \frac{1}{\sigma \sqrt{2\pi}} e ^ {-\frac{(y_i - \hat{y_i})^2}{2\sigma^2}} + \sum_{j = 0}^N \log \frac{1}{\tau \sqrt{2 \pi}} e ^{-\frac{\sum_S (w_j \mathbbm{1}(\mathbb{P}_s (w_j) > T) )^2}{2\tau^2}} \right]\]

\begin{equation}
\mathbb{L}(W) = \argmin_W \left[ \sum_{i=0}^{n} (y_i - (w_0 x_{i, 0} + . . . + w_N x_{i, N} ))^2 + \lambda \sum_{j=0}^{N}\sum_S (w_j \mathbbm{1}(\mathbb{P}_s (w_j) > T) )^2 \right]
\label{loss_expression}
\end{equation}

In all the above equations $\sigma$ and $\tau$ refer to the standard deviation of the prediction and parameters respectively. $\mathbb{P}_s(w_j)$ denotes the probability of weight $w_j$ getting sampled in an $s^{th}$ experiment. $S$ is the set of experiments over which the summation is taken and $T$ is a hyper-parameter used to threshold the probability distribution $\mathbb{P}_s(w_j)$, such that in experiment $s$ weight $w_j$ with $\mathbb{P}_s(w_j) > T$ are selected in the projected space. Following the rigorously proven schema of maximum a-priori estimation, we arrive at the final formulation of the loss function (Eqn. \ref{loss_expression}) which is passed onto to the optimization algorithm to optimize this function over the parameter space $W$.  

\begin{equation}
\mathbb{P}_s (w_j) = \frac{e^{-p_S w_j}}{\sum_{j=0}^{N} e^{-p_s w_j}}
\label{weight_distribution}
\end{equation}

where $p_s \sim \mathbb{U}(0, 1)$,
As the shift in the probability distribution of the parameter space is dynamic, we dampen its effect by introducing the notion of momentum, which is in turn parameterized by a variable $\alpha$. The introduction of momentum reduces the inter-batch variances which results in the smoothening of the loss manifold. 

\begin{equation}
\mathbb{P}_s(w_j) = \alpha \mathbb{P}_s(w_j) + (1 - \alpha) \mathbb{P}_s(w_j^{old})
\label{momentum_prior}
\end{equation}

\subsection{Convexity and Projection Intuition}

The proposed metric can be treated as a special case of group sparsity where the proposed groups are projections of the parameter vectors in a lower-dimensional space. 

\begin{figure}[h]
    \centering
    \subfloat[Sampling Density $s_p$ = 0.01]{\includegraphics[width=0.33\textwidth]{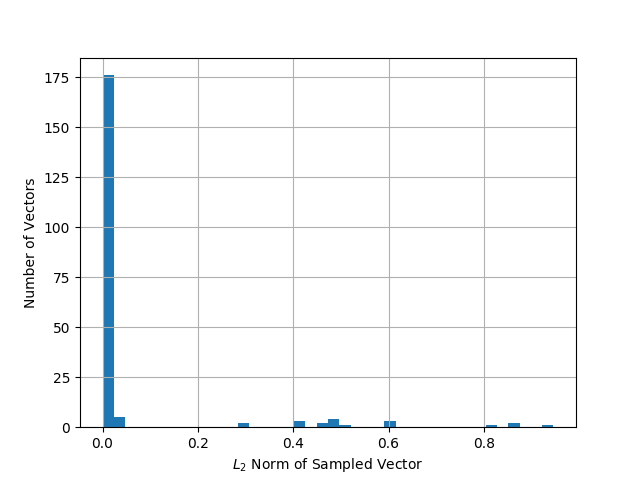}}\hfill
    \subfloat[Sampling Density $s_p$ = 0.05]{\includegraphics[width=0.33\textwidth]{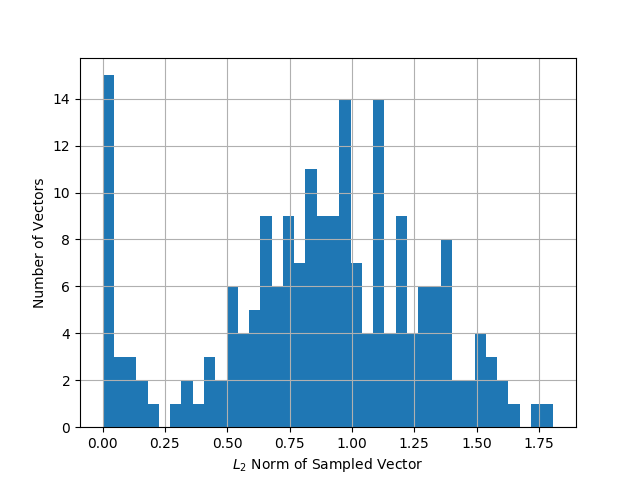}}\hfill
    \subfloat[Sampling Density $s_p$ = 0.1]{\includegraphics[width=0.33\textwidth]{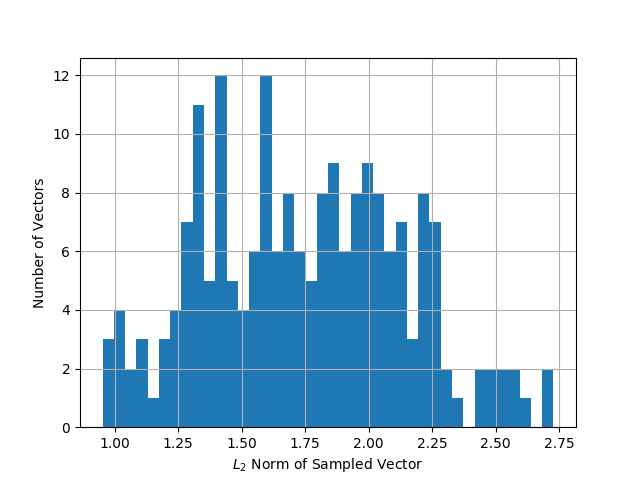}}\hfill
    \caption{Distribution of the $L_2$ norm of the sampled vectors}
    \label{hist_exp}
\end{figure}

The overall heuristic of the process is as follows:-
we reduce the dimension of the parameter vector by projecting it onto a lower-dimensional space. This reduction is done by sampling the axes which are controlled by $s_p$(1-Sparsity or number of ones in a sampled vector), and it reduces the dimensionality drastically. The choice of axes is based on a prior probability distribution and in effect allows parameter points with high magnitude to be present in the \textit{sample vector} as expressed in the Eqn. \ref{weight_distribution}. Now consider the combined effect of sparse sampling with a sparse parent vector. The resulting sampled vectors are majorly of two types: Vectors with high magnitude entries which have a high $L_2$ norm, and vectors with low or zero magnitude entries which have a small or negligible $L_2$ norm as indicated in the Fig. \ref{hist_exp}. When a summation of the $L_2$ norm of all the sampled vectors is done, it ends up overrepresenting significant entries in the parent vector.

\[R(w)_{L1} = \sum_{j=0}^N |w_j| = \sum_{j=0}^N \sqrt{w_j^2}\]
\[R(w)_{L2} = \sqrt {\sum_{j=0}^N w_j ^2} \]
\[R(w)_{proposed} = \lambda \sum_S \sqrt{\sum_{j=0}^{N} (w_j \mathbbm{1}(\mathbb{P}(w_j) > T))^2} = \lambda \sum_S \sqrt{ W^T I_s  I_s^T W}\] where $ I_s $ = $\left[ \mathbbm{1}(\mathbb{P}_s (w_j) > T)\right]_{j=0, 1, 2, ..., N}$\\

Based on the Bayesian framework in Section \ref{method}, the projection of the parameter vector in a lower dimension is done through a thresholded sampling. In an extreme case of sampling where we project the parameter vector on to the set basis vectors composing the space, we end up with the $L_1$ norm of the vector. Numerically, our proposed metric can be compared to a fractional $L_p$ norm, thus proving that the resulting manifold is convex if the parameter vector is convex. While projecting the vector into a lower-dimensional space, the probability of picking one single parameter is higher if the value of that parameter is high. This exercise reduces the effective parameter space to a smaller set that only takes into account significant parameter values. This new smaller parameter space is more significant and leads to an informative and robust prior.

\subsection{Comparison of bounds with $L_1$ and $L_2$ }

\begin{figure}[H]
    \centering
    \includegraphics[width = 0.5\textwidth]{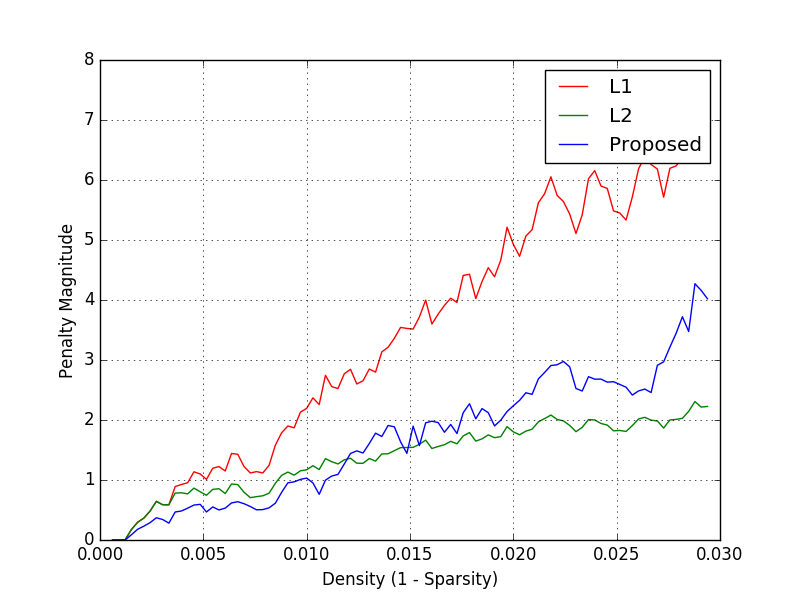}
    \caption{Experiments on bounds estimation}
    \label{penalty_magnitude}
\end{figure}

In this section, we provide a quantitative comparison of different kinds of penalties: proposed and traditional penalties. Figure \ref{penalty_magnitude} shows the magnitude of different penalties by varying the sparsity of the vector on which these penalties act. We are particularly interested in the region where the density (1 - sparsity) of the vector is around 1 \%. In this region, the manifold generated by the proposed metric is constantly below the traditional penalties. Hence, it becomes possible to obtain a better optima. This would theoretically force the optimization function of the parameter to converge on the loss manifold at different optima.

The square root in Eq.\ref{proposed_reg} is the $L_2$ norm of the vector projected on the hyper-plane in a lower-dimensional space. We sample from an experimental space $S$. In this section, we take a simplified version of our proposed metric by using a thresholded uniform prior for the sampling. Here we do not include the momentum factor, thereby making it feasible to study and compare the bounds. We also keep the count of how many times one specific dimension of the parent vector (the vector from which the sampling is done, in this case, $W$) has been sampled.  $\lambda$ and $\lambda '$ are inversely proportional to the $L_{\infty}$ norm of the index counter as obtained from the probability distribution.

\begin{equation}
            R(w)_{proposed} = \lambda \sum_S \sqrt{\sum_{j=0}^{N} (w_j \mathbbm{1}(\mathbb{P}(w_j) > T))^2}
            \label{proposed_reg}
\end{equation}

Taking the expectation over all the experiments and then using Cauchy-Schwarz or Jensen's inequality ($\mathbb{E}(\sqrt x) \leq \sqrt{\mathbb{E}(x)}$). We then arrive at the following inequality:-

\[= \lambda' \mathbb{E}_{\mathbb{P} \sim \mathbb{U}(0, 1)}\left(\sqrt{\sum_{j=0}^N w_j\mathbbm{1}(\mathbb{P}(w_j) > T)^2}\right)\]
\[\leq \lambda' \sqrt{\mathbb{E} \left (\sum_{j=0}^N (w_j\mathbbm{1}(\mathbb{P}(w_j) > T) ^2\right)}\]

Now we try to find an expression for this expectation over $\mathbb{P}(w_j) \sim \mathbb{U}(0, 1) $ in the experiment space $S$ with sparsity $s_p$ constrained on weight $W$s. Here $S$ indicates number of experiments in the Expectation and $N$ indicates the size of the weight vector $W$.

\[ = \lambda' \sqrt{\sum_{j=0}^N w_j^2\mathbb{E} (\mathbbm{1}(\mathbb{P}(w_j) > T)) } \]
\[= \lambda' \sqrt{\sum_{j=0}^N w_j^2 \frac{s_p S (1-T)}{N}}\]
\[=\lambda' \sqrt{\frac{s_p S (1 -T)}{N}} R_{L2}(w)\]

\begin{equation}
    R (w)_{proposed} \leq \lambda' \sqrt{\frac{s_p S (1 -T)}{N}} R(w)_{L2}    
    \label{bound_eq}
\end{equation}

The above Eqn. \ref{bound_eq} proves that for any vector size $N$ and an adequately sampled projections $S$, the proposed penalty is upper bounded by the $L_2$ norm of the same vector. This condition holds within a significant range of $s_p$ and carefully selected experimental space $S$ and threshold $T$.

\subsection{Data}
We illustrate the performance on multiple datasets with various hyper-parameter settings. We make use of MNIST handwritten dataset \cite{mnist_dataset} ( MNIST is a benchmark dataset for images of segmented handwritten digits, containing 28x28 pixel images. It includes 50,000 training examples and 10,000 testing examples), CIFAR-10 and CIFAR-100 datasets \cite{cifar_dataset} (CIFAR-10 and CIFAR-100 dataset consists images of natural images of 32x32 pixels with 10 and 100 classes respectively. CIFAR-10 consists of 60000 color images, with 6000 images per class, 50000 for training and 10000 for testing. CIFAR-100 consists of 60000, 32x32 color images, with 600 images. 50000 for training and 10000 for testing).

\subsection{Experiments}

 We conduct numerous experiments for testing the effectiveness of our proposed method. We train popular networks, with varied sizes of the parameter vector, on MNIST and CIFAR10 datasets by varying the regularization and sparsity. We also perform a class of experiments where we test the proposed metric as a Loss criterion and not just as a regularizer.  Table \ref{exp_tabel} shows the list of experiments done to illustrate the validity of our algorithm. Algorithm \ref{alg:reg_algo} provides the mathematical description of our method.

\begin{table}[H]
\caption{List of Experiments}
\label{exp_tabel}
\begin{tabular}{@{}ccccc@{}}
\toprule
\textbf{Dataset} & \textbf{Regularization} & \textbf{Network}                                                                            & \textbf\textbf{$s_p$}       & \textbf{Loss Type}   \\ \midrule
MNIST            & \{$L_1$, $L_2$, proposed\}    & 2 Layered CNN                                                                               & \{0.01, 0.05, 0.1\} & CE                   \\
CIFAR10          & \{$L_1$, $L_2$, proposed\}    & \begin{tabular}[c]{@{}c@{}}\{2 Layered CNN, \\ VGG-11, ResNet-18,\\ SENet-18\}\end{tabular} & \{0.01, 0.05, 0.1\} & CE                   \\
CIFAR100         & $L_2$                      &  \begin{tabular}[c]{@{}c@{}}\{2 Layered CNN, \\ VGG-11, SENet-18\}\end{tabular}                                                         & --                  & \{CE, Projected CE\} \\ \bottomrule
\end{tabular}
\end{table}

\subsubsection{Training}
Our method was tested with custom convolutional neural networks (CNN) \cite{CNN} with 2 hidden layers along with standard deep networks like VGG-11 \cite{vgg}, ResNet18 \cite{resnet} and DenseNet-121\cite{densenet}. In all the cases the training was performed using Adam \cite{adam} optimizer, with learning rate of 0.001 with batchsize of 32.  

\begin{algorithm}[h]
\caption{Proposed Regularization Method} 
\label{alg:reg_algo}
    \begin{algorithmic}[1]
        \State \textbf{Initialize} $s_p, W, W_{old}, S, T, Reg = 0$
        \For  {$i = 0$ \textbf{to} $S$}
        \State  \quad $\mathbb{P'}(w) = \left (\alpha \frac{e^{-\mathbb{P}(w) w}}{\sum_{j=0}^{N} e^{-\mathbb{P}(w) w}} + ( 1- \alpha)  \frac{e^{- \mathbb{P}(w) w_{old}}}{\sum_{j=0}^{N} e^{-\mathbb{P}(w) w_{old}}}\right )$ \\ 
                       $ \quad \quad \quad \quad \quad \quad \exists \quad \mathbb{P}(w) \sim \mathbb{U} (0,1)  \forall (w, w_{old}) \in (W, W_{old})$
        \State \quad $I_s = [\mathbb{P'}(w_i) > T]_{i=0, 1, 2...,N}$
        \State \quad $W_{projected} = W\odot I_s$
        \State \quad $Reg += ||W_{projected}||^2_2$
        \EndFor
        \State \textbf{end for}
    \end{algorithmic}
\end{algorithm}

\subsubsection{Projected Cross Entropy}
Using the function (\ref{proposed_reg}) defined above for the calculation of the proposed regularizer we use it to revise the cross-entropy loss. The thresholded probability distribution governs the selection of softmax-output. Activations of the final layer replace the roll of weights in the definition of the probability distribution (\ref{weight_distribution}). Performance of a machine learning task with sparse output is greatly affected by the true positives. A projected cross-entropy will improve the performance of the neural network trained on sparse output as it drives the true positives through its focused attention on those axes which have higher activations. 

\subsubsection{Evaluation}
Evaluation of the proposed method was done by comparing the Sparsity of the parameters (not to be confused with $s_p$ which is the sampling sparsity during sampling from the parameter vector), the magnitude of the parameters, overall accuracy and the loss obtained for each training iteration. Each plot provides a comparison between different type of regularizer for a particular metric, where the different types of metrics are loss, accuracy, magnitude, and sparsity.

\textbf{Weight Magnitude and Sparsity :}  We choose sparsity and magnitude as an evaluation metric for the following reasons:
First, regularizers have been conventionally linked to the sparsity of parameters while $L_1$ norm induces direct sparsity. Further, $L_2$ norm induces a continuity and boundedness on the parameters. Both these regularizers, in turn, have an effect of reducing the magnitude of the parameters. 
Second, sparsity plays a major role in the interpretability and pruning of a network. While we make no claims in this paper about these two ideas, we do report how this metric changes when the proposed metric changes in comparison with $L_2$ norm.
\[ Sparsity = 1 - \frac{\sum_{i=0}^N \mathbbm{1}(w_i > T) }{N} \]
\textbf{Accuracy and Loss:} Both these metrics are of the utmost importance for proposing changes to any part of the deep learning regime, be it a new loss or a new architecture. A lower testing loss indicates that the optimization algorithm has found a better optimum in the loss manifold. By extension, a higher value of testing accuracy shows that the point in parameter space corresponding to the optimum leads to a better network configuration which is superior at generalizing. This is a highly desirable quality in a network of any class.

\section{Results and Discussion}

\begin{table}[h]
\centering
\caption{Results of proposed algorithm on various datasets and models}
\label{results_tabel}
\begin{tabular}{@{}cccccc@{}}
\toprule
\textbf{Dataset}           & \textbf{Network}                                & \textbf{Loss}                              & \textbf{Reg}                    & \textbf{Wt. Density}                         & \textbf{Accuracy} \\ \midrule
\multirow{2}{*}{MNIST-10}  & \multirow{2}{*}{2 Layered CNN}                  & \multirow{2}{*}{CE}                        & \textbf{Proposed}               & \textbf{1.2 x $10^{-3}$}                      & \textbf{0.992}    \\
                           &                                                 &                                            & $L_2$                            & 1.6 x $10^{-3}$                               & 0.990             \\           &                                &              \\ 
                           \midrule
\multirow{12}{*}{CIFAR-10}  & \multirow{1}{*}{2 Layered CNN}                  & CE                                         & \textbf{Proposed}               & 2.4 x $10^{-3}$                      & \textbf{0.422}    \\
                           &                                                 &                                            & $L_2$                            & 3.2 x $10^{-3}$                               & 0.38              \\
                           &                                                 &                                            & $L_1$                            & \textbf{1.8 x $10^{-3}$}                              &  0.16             \\
                           
                           & \multirow{1}{*}{VGG-11}                         & CE                                         & \textbf{Proposed}               & 5.5 x $10^{-4}$                      & \textbf{0.883}    \\
                           &                                                 &                                            & $L_2$                            & 6.3 x $10^{-4}$                               & 0.875             \\
                           &                                                 &                                            & $L_1$                            & \textbf{4.8 x $10^{-4}$}                               & 0.64              \\
                           
                           & \multirow{1}{*}{ResNet-18}                      & CE                                         & \textbf{Proposed}               & \textbf{8.3 x $10^{-4}$}                      & \textbf{0.910}    \\
                           &                                                 &                                            & $L_2$                            & 9.1 x $10^{-4}$                               & 0.908             \\
                           &                                                 &                                            & $L_1$                            & 8.6 x $10^{-4}$                               & 0.862              \\
                           
                           & \multirow{1}{*}{SENet-18}                       & CE                                         & \textbf{Proposed}                        & 4.2 x $10^{-4}$                               & \textbf{0.913}             \\
                           &                                                 &                                            & $L_2$                            & 3.9 x $10^{-4}$                               & 0.910              \\
                           &                                                 &                                            & $L_1$                            &\textbf{4.0 x $10^{-4}$}                             & 0.843              \\
                           
                           \midrule
\multirow{4}{*}{CIFAR-100} & \multirow{1}{*}{VGG-11}                         & \textbf{Projected CE}                      & -                               & \textbf{4.2 x $10^{-4}$}                      & \textbf{0.59}     \\
                           &                                                 & CE                                         & -                               & 5.5 x $10^{-4}$                               & 0.57              \\
                           & \multirow{1}{*}{ResNet-18}                      & \textbf{Projected CE}                      & -                               & \textbf{5.3 x $10^{-4}$}                      & \textbf{0.60}     \\ 
                           &                                              & CE                                         & -                               & 8.5 x $10^{-4}$                     & 0.59                                       \\ \bottomrule
\end{tabular}
\end{table}

\subsection{Proposed Regularization Results}
\begin{figure}[H]
    \centering
    \subfloat[Loss]{\includegraphics[width=0.25\textwidth]{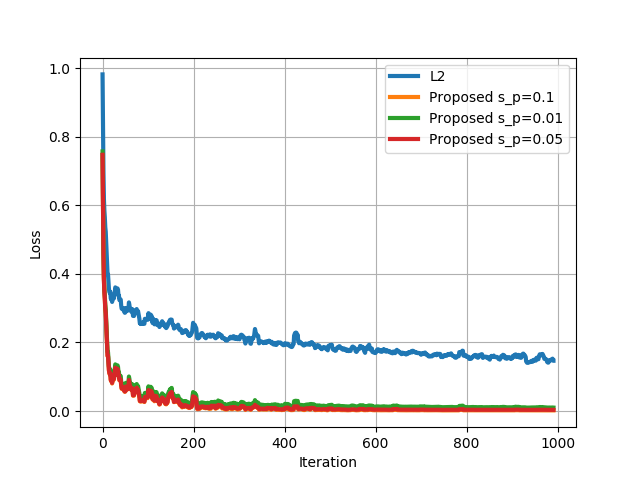}}\hfill
    \subfloat[Accuracy]{\includegraphics[width=0.25\textwidth]{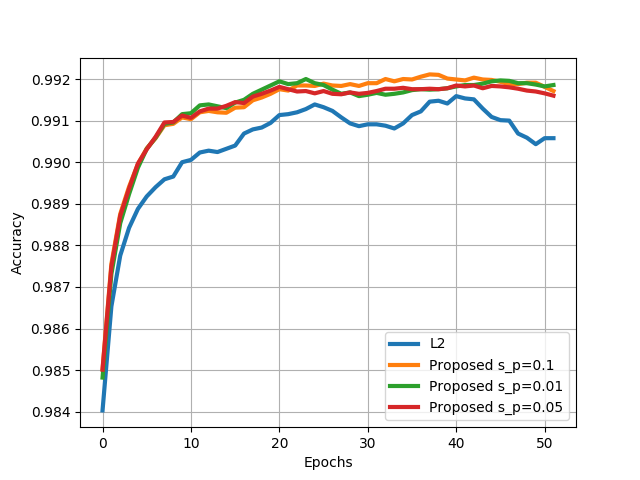}}\hfill
    \subfloat[wt. Magnitude]{\includegraphics[width=0.25\textwidth]{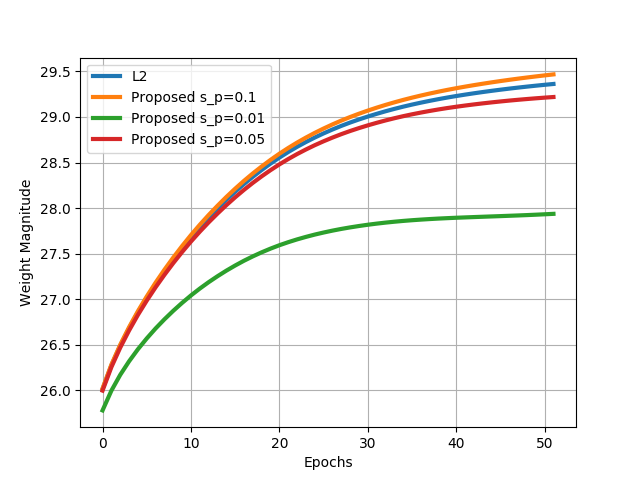}}\hfill
    \subfloat[wt. Density]{\includegraphics[width=0.25\textwidth]{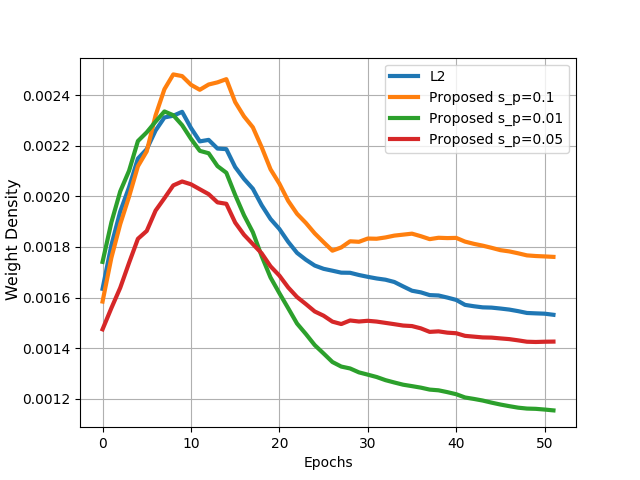}}\hfill
    \caption{The above plot shows the obtained  (a) loss, (b) accuracy, (c) wt. magnitude and (d) wt. density for 2 layered CNN trained on MNIST dataset }
    \label{mnist_results}
\end{figure}

\begin{figure}[H]
    \centering
    \subfloat{\includegraphics[width=0.25\textwidth]{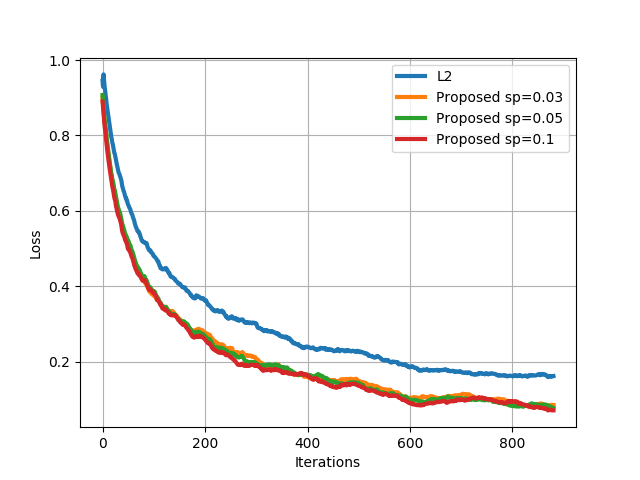}}\hfill
    \subfloat{\includegraphics[width=0.25\textwidth]{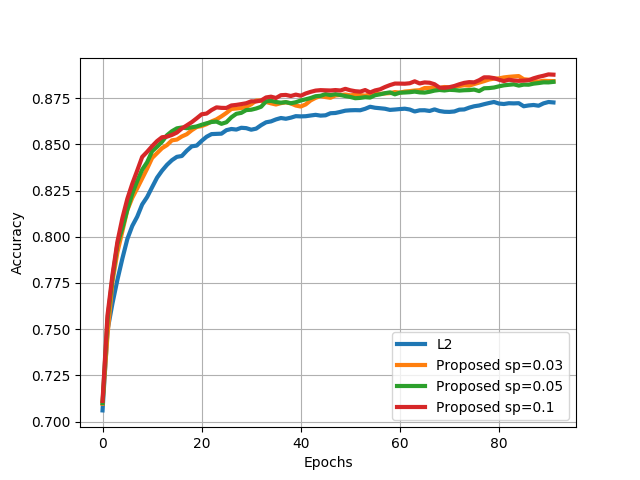}}\hfill
    \subfloat{\includegraphics[width=0.25\textwidth]{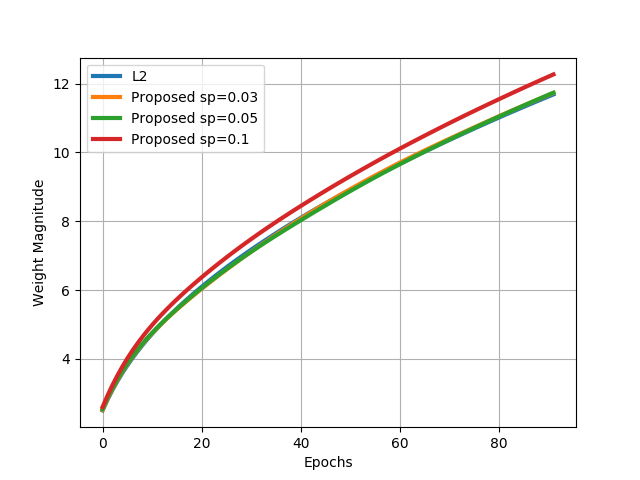}}\hfill
    \subfloat{\includegraphics[width=0.25\textwidth]{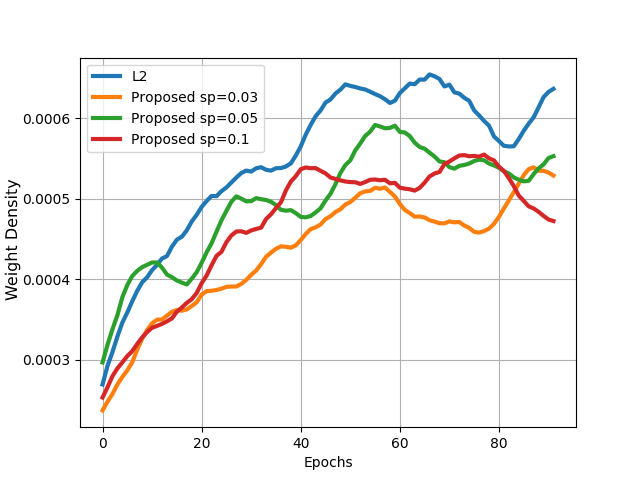}}\hfill
    \caption{The above plot shows the obtained  (a) loss, (b) accuracy, (c) wt. magnitude and (d) wt. density for SENet-18 trained on CIFAR-10 dataset }
    \label{cifar_results}
\end{figure}

In each plot illustrated in this section, different colors signify experiments done by varying the sampling sparsity $s_p$. With respect to Loss and Accuracy, the proposed method clearly outlies traditional $L_2$ norm (Fig. \ref{mnist_results} \& Fig. \ref{cifar_results}) . The idea is that the addition of proposed regularization penalty shifts the manifold of the testing loss and makes the parameters to converge at a point which has a higher testing accuracy, thus pushing the model toward a better generalization domain.

For the induced sparsity and magnitude of the parameters, one configuration ($s_p = 0.01$) of the proposed method stands apart from the rest. The combined magnitude of the parameters differs for the proposed metric. This change leads to a parameter configuration which is both lower in magnitude as well as sparse in density. This might lead to a better analysis of the interpretability of the model in future works. Here we have used the notion of momentum in the probability distribution in all the experiments as described in section \ref{momentum_prior}. Table \ref{results_tabel} shows the effect of the proposed regression on MNIST and CIFAR-10 data on various models with $s_p = 0.01$. Figure \ref{mnist_results} and \ref{cifar_results} shows the plots of our method with multiple $s_p$ values and standard $L_2$ regularizer with MNIST and CIFAR dataset respectively.



\subsection{Proposed loss results}
\begin{figure}[H]
    \centering
    \subfloat{\includegraphics[width=0.25\textwidth]{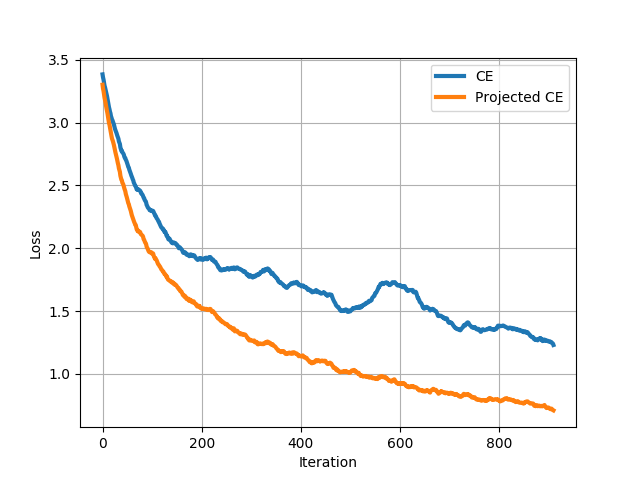}}\hfill
    \subfloat{\includegraphics[width=0.25\textwidth]{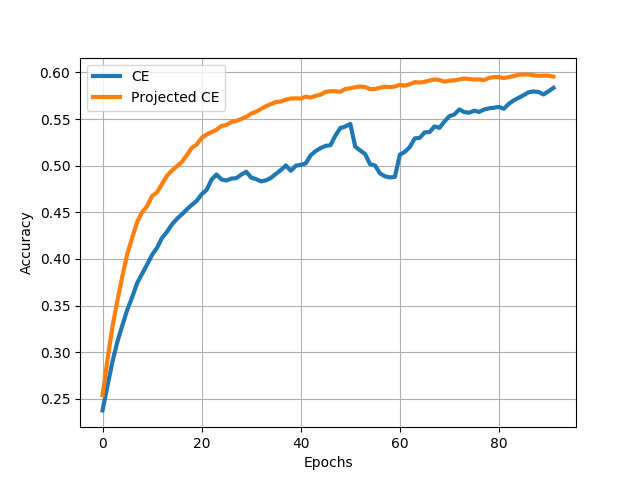}}\hfill
    \subfloat{\includegraphics[width=0.25\textwidth]{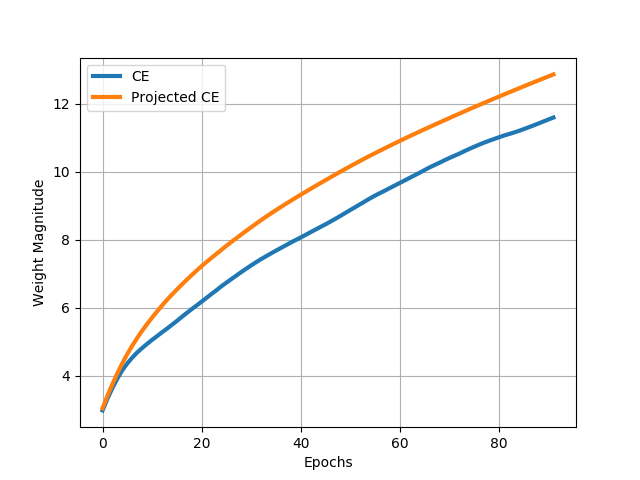}}\hfill
    \subfloat{\includegraphics[width=0.25\textwidth]{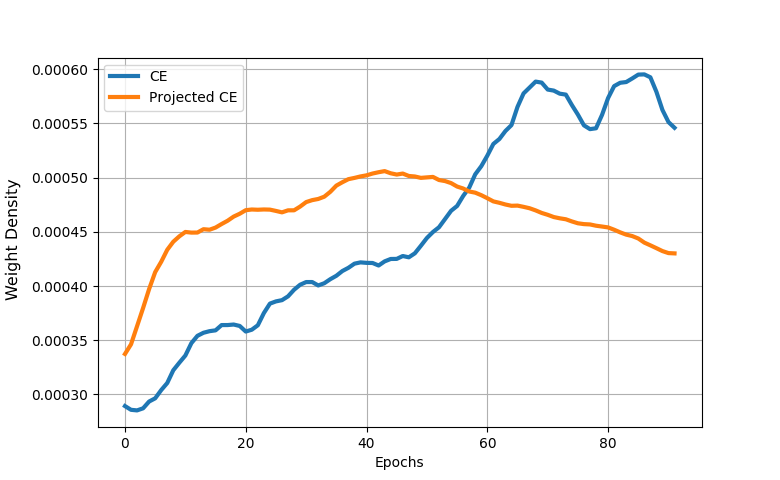}}\hfill
    \caption{The above plot shows the obtained  (a) loss, (b) accuracy, (c) wt. magnitude and (d) wt. density for SENET-18 trained on CIFAR-100 dataset }
    \label{cifar100_results}
\end{figure}

Table \ref{results_tabel} shows the effect of the projected loss on CIFAR-100 data. Figure \ref{cifar100_results} shows the plots of our method (projected loss) with CIFAR-100 dataset. The training regime of a deep neural network in a traditional setting follows the following arc. During the initial few iterations the test loss falls without kinks and the test accuracy also increases in a similar manner (around 200 iterations or 20 epochs in this case). When the training proceeds further, since the loss is calculated over the entire output layer, the gradient received at each weight is comparatively stronger than required. This leads to erratic or oscillatory behavior. When a projected version of the same loss is used, the nodes whose magnitude is significantly higher than the rest are taken into consideration, leading to an appropriate backflow of error. This can also be the reason for a much smoother accuracy and loss curve. Classification problems with multiple classes are clear targets for the projected loss because for one data-point, there are only a few nodes which should be activated and doing a piece-wise projected loss ensures such a behavior.

\section{Future Work}
Though we find better results in the initial study, there still remains a lot of experimentation and analyses to be done. One direction is to explore datasets which can benefit from these changes, to further explore in this direction, we would like to test the effect of proposed method over sparse NLP datasets. An additional line of study would be to explore the resulting sparse networks through the lens of compression and conduct the tests for interpretability of the obtained network.

\section{Conclusion}
In this work, we propose different regularizers, which works on the projected spaces of the original vector. Our approach makes use of the current state of the parameters in the network which results in informative and dynamically varying prior for regularization. Our experiments on MNIST, CIFAR-10, and CIFAR-100 with multiple custom and pre-existing networks like VGG, ResNet, and SENet, illustrate the better results in terms of accuracy, loss, magnitude, and sparsity when experimented with the proposed regularizer as compared to the standard $L_2$ regularizer.

\bibliographystyle{unsrt}
\bibliography{ref.bib}

\end{document}